\documentclass[11pt,a4paper,table]{article}
\usepackage[hyperref]{emnlp2020}
\usepackage{times}
\usepackage{latexsym}

\usepackage[utf8]{inputenc}
\usepackage{tabularx,graphicx}
\usepackage{amsthm}
\usepackage{soul}
\usepackage{booktabs} 
\usepackage{subcaption}
\usepackage{amssymb}
\usepackage{multirow}
\usepackage{xcolor}
\usepackage{pifont}% http://ctan.org/pkg/pifont
\newcommand{\cmark}{\ding{51}}%
\newcommand{\xmark}{\ding{55}}%

\aclfinaltrue

\title{Does Data Augmentation Improve Generalization in NLP?}

\author{
   Rohan Jha \\
  Brown University\\
  \And 
   Charles Lovering \\
  Brown University\\
  \{\texttt{first}\}\texttt{\_}\{\texttt{last}\}\texttt{@brown.edu} \\
  \And
  Ellie Pavlick \\
  Brown University\\
 }

\begin{document}
\maketitle
\begin{abstract}
Neural models often exploit superficial features to achieve good performance, rather than deriving more general features. Overcoming this tendency is a central challenge in areas such as representation learning and ML fairness. Recent work has proposed using data augmentation, i.e., %\charliereplace{generating training examples on which the dispreferred features fail}{
generating training examples where the superficial features fail, as a means of encouraging models to prefer the stronger features. We design a series of toy learning problems to test the hypothesis that data augmentation leads models to unlearn weaker heuristics,
but not to learn stronger features in their place.
%\charliereplace{but not to prefer}{rather than learn the} stronger features \charliereplace{in their place}{instead}.
We find partial support for this hypothesis: Data augmentation often hurts before it helps, and it is less effective when the preferred strong feature is much more difficult to extract than the competing weak feature.  
\end{abstract}

\section{Introduction}
Neural models often perform well on tasks by using superficial (``weak'') features, rather than the more general (``strong'') features that we'd prefer them to use. As a result, models often fail in systematic ways. For example, in visual question answering, models failed when tested on rare color descriptions (\textit{``green bananas''}) \citep{vqa-cp}; in coreference, models failed when tested on infrequent profession-gender pairings (\textit{``the nurse cared for his patients''}) \citep{rudinger-etal-2018-gender}; in natural language inference (NLI), models failed on sentence pairs with high lexical overlap but different meanings (\textit{``man bites dog''}/\textit{``dog bites man''}) \citep{mccoy2019right}. One proposed solution has been to augment training data with ``counterexamples'' to the model's adopted heuristics, i.e., training examples on which the weak features fail. This technique has shown positive initial results for POS tagging \cite{elkahky-etal-2018-challenge}, NLI \cite{min2020augmentation}, and reducing gender bias \cite{zhao-etal-2018-gender,zmigrod-etal-2019-counterfactual}. However, it is not yet known whether this strategy is a feasible way of improving systems in general.

Our hypothesis is that augmenting data with such counterexamples will lead models to unlearn weaker features as intended (e.g., reducing the correlation between gender and profession) but will not necessarily lead models to adopt the ``better'' stronger features we intend them to use instead (e.g., syntax, common sense inference). To test this, we design a set of toy learning problems that contain two competing features which differ both in how well they predict the label and in how easy they are for the model to extract.
%We analyze model performance as a function of task difficulty and dataset skew.
We find that:
\begin{itemize}
\item Data augmentation is less effective when the preferred (``strong'') feature is harder for the model to extract from its input (\S\ref{sec:hardness}). %This suggests that improving input representations is likely to be as, possibly more, effective than data augmentation at leading the model to use the stronger feature.
%\charliequestion{This seems like a claim we can't substantiate.} \elliecomment{will move to discussion}
\item Adding counterexamples often hurts before it helps, and at low levels ($\sim\leq1\%$ of training data), data augmentation leads models to shift to new weak features rather than to the stronger feature (\S\ref{sec:heuristics}). This suggests apparent gains from data augmentation could be misleading, especially if we only evaluate on the targeted phenomenon.
\item Data augmentation only works when the strong feature is already sufficiently present in training. When training data is pathologically skewed such that the strong feature very rarely or never occurs without the weak feature, data augmentation has no effect (\S\ref{sec:skew}).
\end{itemize}

\paragraph{Note on Terminology:}

The goal of this study is to isolate the phenomenon of interest (data augmentation in NLP) and observe patterns without the confounds that exist in applied settings. By homing in on specific empirical trends, we hope to lay the groundwork for more formal subsequent analysis. As such, our choice of terminology (\textit{``strong''}, \textit{``weak''}, \textit{``hard''}, \textit{``counterexample''}) is meant to be informal. We deliberately avoid more widely-used terms (\textit{``adversarial examples''}, \textit{``spurious correlations''}, \textit{``counterfactual''}, etc.) since we do not yet wish to invoke specific connotations or assumptions about modeling approach, causality, or downstream application.

\section{Experimental Design}

%\subsection{Intuition}

%Our study is motivated by two empirical findings presented in \citet{mccoy2019right}. Specifically, \citet{mccoy2019right} focused on models' use of syntactic heuristics in the context of the natural language inference (NLI) task: given a pair of sentences--the premise $p$ and the hypothesis $h$--predict whether or not $p$ entails $h$. They showed that when 1\% of the 300K sentence pairs seen in training exhibit lexical overlap (i.e. every word in $h$ appears in $p$) and 90\% of lexical-overlap sentence pairs have the label \textsc{entailment}, the model adopts the (incorrect) heuristic that lexical overlap always corresponds to \textsc{entailment}. However, after augmenting the training data with automatically generated training examples so that 10\% of the 300K training pairs exhibit lexical overlap and 50\% of lexical-overlap sentence pairs have the label \textsc{entailment}, the same model did \textit{not} adopt the heuristic and appeared to learn features which generalized to an out-of-domain test distribution. 

%From these results, it is hard to say which changes to the training setup were most important for the model's improved generalizability. The number of lexical-overlap examples seen in training? The probability of \textsc{entailment} given that a pair exhibits lexical overlap? Or some other positive artifact of the additional training examples? Thus, we abstract away from the specifics of the NLI task in order to consider a simplified setting that captures the same intuition but allows us to answer such questions more precisely. 

\subsection{Setup}
\begin{figure}[ht!]
\centering
\includegraphics[width=.9\linewidth]{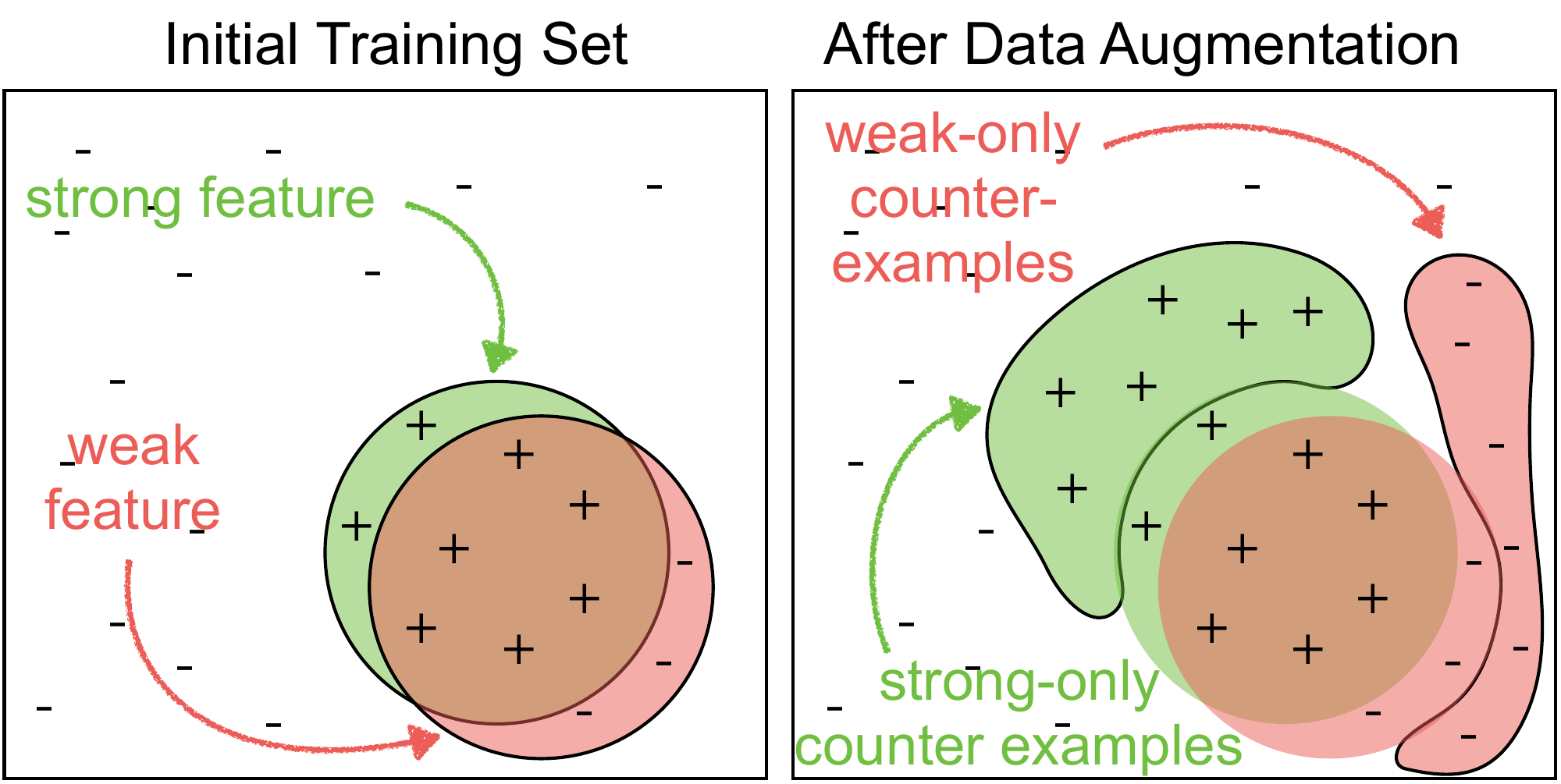}
\caption{Schematic of experimental setup.}
\label{fig:error-cases}
\end{figure}

We use a binary sequence classification task. We assume there exists some \textbf{strong feature} which perfectly predicts the label (i.e., the label is $1$ \textit{iff} the strong feature holds), but which is non-trivial to extract given the raw input. Additionally, there exists a \textbf{weak feature} which is easy for the model to extract from the input and which frequently co-occurs with the strong feature in training. Thus, a model which only represents the weak feature can make correct predictions much of the time. We can vary the features' co-occurence by adding \textbf{counterexamples} to training in which either the strong or the weak feature is present, but not both (see Fig.\ \ref{fig:error-cases}). Intuitively, we intend the weak feature to be representative of features such as lexical priors (e.g., \textit{``not''} being indicative of contradiction in NLI) and the strong feature to be representative of syntactic and compositional semantic features. %\cut{derivable from a sentence}.

%\footnote{See Appendix \ref{app:noise} for results which assume varying levels of label noise.}

\subsection{Task and Model}
We use a synthetic sentence classification task with $k$-length sequences of numbers as input and binary labels as output. We use a symbolic vocabulary $V$ with the integers $0 \dots |V|$. We fix $k=10$ and $|V|=50$K. We use an initial training size of $200$K examples, though the total training size varies as we add counterexamples (\S\ref{sec:counterexamples}). Test and validation sizes are $40$K each.
%We do see some effects associated with vocabulary size, but none that affect our primary conclusions; see Appendix \ref{app:vocab} for details.
Our classifier is a simple network with an embedding layer, a 1-layer LSTM, and an MLP with 1 hidden layer and ReLU activation (about 13M parameters). We use Intel Cascade CPUs, and models are trained until convergence using early-stopping. Various ablations are included in Supplementary Material (\S \ref{sec:ablations}).
The test and validation vocabularies are disjoint from the training; for the model to test well, it has to learn the intended strong feature (i.e., the notion of a duplicated symbol) vs.\ finding an easier workaround (e.g., memorizing bigrams). Models converge on average by 8 epochs (500 seconds each). Our code is available for reproducibility.\footnote{
Code at
\url{http://bit.ly/bb-data-aug}.}
% \rohan{Also includes code to compute the MDL and the MI heatmaps.}
% \rohan{Experiments (with MDL calculations; \S\ref{sec:strong-weak-hardness}) take YYY on average on ZZZ.} 

%  & $5.0\times10^{-1}$
% & $9.6\times10^{-1}$
%  &  $1.8\times10^{0}$ 
% & $1.0\times10^{0}$
\begin{table*}[ht!]
\centering
\resizebox{\linewidth}{!}{
\begin{tabular}{lllll}
\toprule
Feature Nickname & Description & Data MDL & Model MDL & Example\\
\midrule
\texttt{contains-1} & \texttt{1} occurs somewhere in the sequence & $1.2\times10^{1}$ & $2.8\times10^{2}$ & \texttt{2 4 11 \colorbox{gray}{1} 4} \\
\texttt{prefix-dup} & Sequence begins with a duplicate  & $3.0\times10^{2}$ & $8.0\times10^{4}$ & \texttt{\colorbox{gray}{2 2} 11 12 4} \\ %$0.517$ 
\texttt{first-last} & First number equals last number & $3.3\times10^{2}$ & $1.2\times10^{5}$ & \texttt{\colorbox{gray}{2} 4 11 12 \colorbox{gray}{2}} \\ %$0.546$
\texttt{adjacent-dupl}  & Adjacent duplicate anywhere in seq.  & $5.8\times10^{2}$ & $2.5\times10^{5}$ & \texttt{11 12 \colorbox{gray}{2 2} 4} \\ %$1.429$
%\texttt{contains-first}  & First number is elsewhere in seq. & $3.0\times10^{0}$ & $8.4\times10^{3}$ & $4.7\times10^{2}$ & \texttt{\colorbox{gray}{2} 11 \colorbox{gray}{2} 12 4}\\ %$1.542$
%\texttt{contains-1} & \texttt{1} occurs somewhere in the sequence & 0.17 & 0.08 & 0.294 & \texttt{2 4 11 \colorbox{gray}{1} 4} \\
%\texttt{prefix-dup} & Sequence begins with a duplicate &  0.70 & 160 & 153 & \texttt{\colorbox{gray}{2 2} 11 12 4} \\ %$0.517$ 
%\texttt{first-last} & First number equals last number &  1.29 & 460 & 232 & \texttt{\colorbox{gray}{2} 4 11 12 \colorbox{gray}{2}} \\ %$0.546$
%\texttt{adj-dupl}  & Adjacent duplicate anywhere in seq. & 0.72 & 669 & 214 & \texttt{11 12 \colorbox{gray}{2 2} 4} \\ %$1.429$
%\texttt{contains-first}  & First number is elsewhere in seq. & 3.03 & 8436 & 470 & \texttt{\colorbox{gray}{2} 11 \colorbox{gray}{2} 12 4}\\ %$1.542$
\bottomrule
\end{tabular}}
\caption{Features used to instantiate the strong feature in our experiments. Features are intended to differ in how hard they are for an LSTM to detect given sequential input. The MDL statistics are measured in bits (\S\ref{sec:strong-weak-hardness}).}
\label{tab:true-properties}
\end{table*}

\subsection{Strong and Weak Features}\label{sec:strong-weak-hardness}
In all experiments, we set the weak feature to be the presence of the symbol \texttt{2} anywhere in the input. We consider several different strong features (Table \ref{tab:true-properties}), intended to vary in how ``hard'' they are for the classifier to detect. To quantify the notion of \textbf{hardness}, we use minimum description length (MDL), a recently introduced information-theoretic metric designed to capture the ``extractability'' of a feature from a representation \citep{voita2020information}. Intuitively, MDL computes the minimum number of bits that would be needed to encode the labels given the representations: The more directly the representations encode the labels, the fewer the bits needed. To compute MDL\footnote{We compute the online MDL; see \citep{voita2020information} for implementation details.}, we train a model to predict directly whether each feature holds, using a set of $200$K training examples evenly split between cases when the feature does and does not hold. Table \ref{tab:true-properties} contains the MDL metrics for each feature (averaged over 3 re-runs). We see the desired gradation of feature hardness. The validation error of the converged classifier is $< 7.5$\% in all cases.

\subsection{Performance Metrics}

%\charliereplace{We partition test accuracy into four regions of interest, defined below. We use a finer-grained partition than standard true positive rate and false positive rate because we are particularly interested in measuring accuracy in relation to the presence or absence of the weak feature.}{
We partition test accuracy into four cases,
%, rather than the standard true positive rate and false positive rate, because
since we are interested in measuring accuracy in relation to the presence or absence of the weak feature. 

\begin{tabular}{ll}
\textbf{\textit{weak-only} acc:} & $P(pred=0 \mid	wk,		\neg str)$ \\
\textbf{\textit{strong-only} acc:}  & $P(pred=1      \mid		\neg wk,	str)$ \\
\textbf{\textit{both} acc:}  & $P(pred=1		\mid		wk, str)$ \\
\textbf{\textit{neither} acc:}  & $P(pred=0		\mid		\neg wk,	\neg str)$ \\
\end{tabular}

%\noindent For completeness, we define and compute \textit{both} error and \textit{neither} error. However, in practice, since our toy setting contains no spurious correlations other than those due to the weak feature, we find that these two error metrics are at or near zero for all of our experiments (except for a small number of edge cases discussed in \S\ref{app:case-2-3-error}). Thus, for all results in Section \ref{sec:results}, we only show plots of \textit{strong-only} and \textit{weak-only} error, and leave plots of the others to Appendix \ref{app:case-2-3-error}.

\noindent Our test and validation sets have $10$K examples for each of these cases.\footnote{Validation is always within 1\% of average test error.} Intuitively, low \textit{weak-only} accuracy indicates the model associates the weak feature with the positive label, whereas low \textit{strong-only} accuracy indicates the model either fails to detect the strong feature altogether, or detects it but fails to associate it with positive label. In practice, we might prioritize these accuracies differently in different settings. For example, in work on fairness \citep{hall-maudslay-etal-2019-name}, we target \textit{weak-only} accuracy, since the primary goal is to prevent the model from falsely associating protected attributes with specific labels or outcomes. In contrast, when aiming to improve the overall robustness of NLP models \citep{min2020augmentation}, we presumably target \textit{strong-only} accuracy, since the hope is that, by lessening the effectiveness of shallow heuristics, we will encourage models to learn deeper, more robust features instead. 

\subsection{Counterexamples}
\label{sec:counterexamples}

%Data augmentation aims to reduce the above errors by generating new training examples which decouple the strong and weak features.
As with error categories, we distinguish between types of counterexamples: \textbf{\textit{weak-only} counterexamples} where the weak feature occurs without the strong feature and the label is $0$, and \textbf{\textit{strong-only} counterexamples} where the strong feature occurs without the weak feature and the label is $1$. These two types are meaningfully different in practical settings, as often  \textit{weak-only} counterexamples are easy to obtain, whereas \textit{strong-only} counterexamples are difficult or impossible to obtain. For example, in the case of NLI and the lexical overlap heuristic from \citet{mccoy2019right}, it is easy to generate \textit{weak-only} counterexamples (sentence pairs with high lexical overlap but which are not in an entailment relation) using a set of well-designed syntactic templates. However, generating \textit{strong-only} examples (entailment pairs without lexical overlap) likely requires human effort \citep{multinli}. This difference is exacerbated in realistic problems where there are many weak features and/or it is impossible to fully isolate the strong features from all weak features (e.g., ultimately, it may not be possible to decouple the ``underlying semantics'' of a sentence from all lexical priors). We characterize most recent applied work on data augmentation  \citep{elkahky-etal-2018-challenge,min2020augmentation,zhao-etal-2018-gender,zmigrod-etal-2019-counterfactual} as using \textit{weak-only} examples. 
 
We begin with an initial training set of $200$K, evenly split between \textit{both} and \textit{neither} examples.
We then add all combinations of $i$ \textit{weak-only} counterexamples and $j$ \textit{strong-only} counterexamples for $i,j\in$\{$0$, $10$, $100$, $500$, $1$K, $2.5$K, $5$K, $25$K, $50$K, $100$K\}. This results in 1,191 total models trained: 3 random restarts\footnote{It isn't always exactly 3 restarts due to some job failures.} for 100 combinations of $i, j$ over the four features in Table \ref{tab:true-properties}.

\section{Initial Results}
\label{sec:regression}

Using this terminology, our hypothesis is: Adding \textit{weak-only} counterexamples will improve \textit{weak-only} accuracy but will not improve \textit{strong-only} accuracy. As a first-pass analysis, we use a standard multiple regression.\footnote{\url{linear_model.OLS} from \url{statsmodels}} This allows us to observe the first-order effect of the number and type of counterexample on accuracy, controlling for the effects that additional counterexamples have on total training size and on training label skew. For the moment, we focus only on highlighting general relationships between the number and type of counterexample added and the model's \textit{weak-only} and \textit{strong-only} accuracies. We focus in this way for simplicity, to give a starting point of our analysis, and don't claim that this tells the entire story.
%note that the model's overall performance is dependent on all accuracies together;
In Section \ref{sec:analysis}, we look more deeply at the trends observed in relation to \textit{both} and \textit{neither} accuracy. 

%\subsection{Regression Analysis}

We run four regressions, one per accuracy metric. We include variables to control for the confounding effects of adding counterexamples, e.g., the label skew and the total number of examples. These aren't shown because of space constraints.\footnote{Supplementary has these metrics and \textit{both} and \textit{neither}.} As preprocessing, we take the log of number of counterexamples and of total training size, and we normalize all explanatory variables to have a mean of $0$ and standard deviation of $1$. Table \ref{tab:mains} shows the regression results. There are two main takeaways.

First, we see a significant, positive relationship both between the number of \textit{weak-only} examples and \textit{weak-only} accuracy ($\beta=0.37$, $p=0.00$) but no significant relationship between the number of \textit{weak-only} examples and \textit{strong-only} accuracy ($\beta=0.03$, $p=0.25$). This is consistent with our hypothesis: All else being equal, \textit{weak-only} examples have a much larger effect on \textit{weak-only} accuracy than on \textit{strong-only} accuracy. When looking at the effects associated with adding  \textit{strong-only} examples, we see a strong positive effect on  \textit{strong-only} accuracy ($\beta=0.46$, $p=0.00$), but a modest negative effect on \textit{weak-only} accuracy ($\beta=-0.14$, $p=0.00$). This seems counterintuitive; we investigate further in 
%to stem from U-shaped learning curves, which we will discuss in detail in
Section \ref{sec:heuristics}.

The second trend that stands out from the regression results is the clear association between the hardness of the strong feature, as measured by MDL, and the model's accuracy. That is, harder features have more negative impact on accuracy. To understand this trend better, we run a second regression that includes interaction terms between the number of counterexamples and each specific strong feature. We find that our hardest feature (\texttt{adjacent-dupl}) behaves differently than all the others: \textit{strong-only} examples hurt \textit{weak-only} accuracy ($\beta=-0.09, p=0.00$), and \textit{weak-only} examples fail to help \textit{strong-only} accuracy ($\beta=-0.05, p=0.17$). For all other features, counterexamples have a positive (albeit asymmetric) effect on both metrics: \textit{weak-only} examples have a roughly $3\times$ to $5\times$ larger effect on \textit{weak-only} accuracy than on \textit{strong-only} accuracy, and vice-versa for \textit{strong-only} examples.\footnote{See Supplementary Material (Table \ref{tab:mains-interactions}).}

%\charliequestion{Does strong and significant mean something special here?($\beta=-0.09$) being significant vs ($\beta=0.47$) being strong seems strange to me. Significant sounds like a key word. Perhaps, strong and modest? }
\begin{table}[ht!]
\centering
\small
\setlength{\tabcolsep}{.5em}
\begin{tabular}{lcrlrl}
\toprule
&  & \multicolumn{2}{c}{\textit{weak-only} acc.} & \multicolumn{2}{c}{\textit{strong-only} acc.} \\
&  & \multicolumn{2}{c}{$R^2=0.74$}&\multicolumn{2}{c}{$R^2=0.70$} \\
\cmidrule(lr){3-4} \cmidrule(lr){5-6}
& $\sigma$ & \multicolumn{1}{c}{$\beta$} & \multicolumn{1}{c}{$p$} & \multicolumn{1}{c}{$\beta$} & \multicolumn{1}{c}{$p$} \\
%\cmidrule(lr){2-2} \cmidrule(lr){3-4} \cmidrule(lr){5-6}
\midrule
log weak-only ex. & $1.5$&{$0.37$}&{$0.00$*}&{$0.03$}&{$0.25$} \\
log strong-only ex. & $1.5$&{$-0.14$}&{$0.00$*}&{$0.46$}&{$0.00$*} \\ 
\midrule
contains-1 &    &{$0.89$}&{$0.00$*}&{$0.90$}&{$0.00$*} \\ 
prefix-dupl &    &{$-0.07$}&{$0.03$*}&{$-0.01$}&{$0.71$} \\ 
first-last &    &{$-0.26$}&{$0.00$*}&{$-0.43$}&{$0.00$*} \\ 
adjacent-dupl &    &{$-0.58$}&{$0.00$*}&{$-0.48$}&{$0.00$*} \\ 
\bottomrule
\end{tabular}
\caption{Regression results for predicting \textit{weak-only} and \textit{strong-only} accuracy. All explanatory variables are $z$-normalized, so coefficients should be interpreted in units of $\sigma$ (variable's st. dev.). First section shows effects of number and type of counterexamples. Second section shows effects of dummy variables for each individual feature type.
%in order of increasing hardness (Table \ref{tab:true-properties}). \rohanquestion{I'm somewhat confused about the second section.}
}
\label{tab:mains}
\end{table}

\begin{figure*}[ht!]
    \centering
                  \includegraphics[width=.8\linewidth]{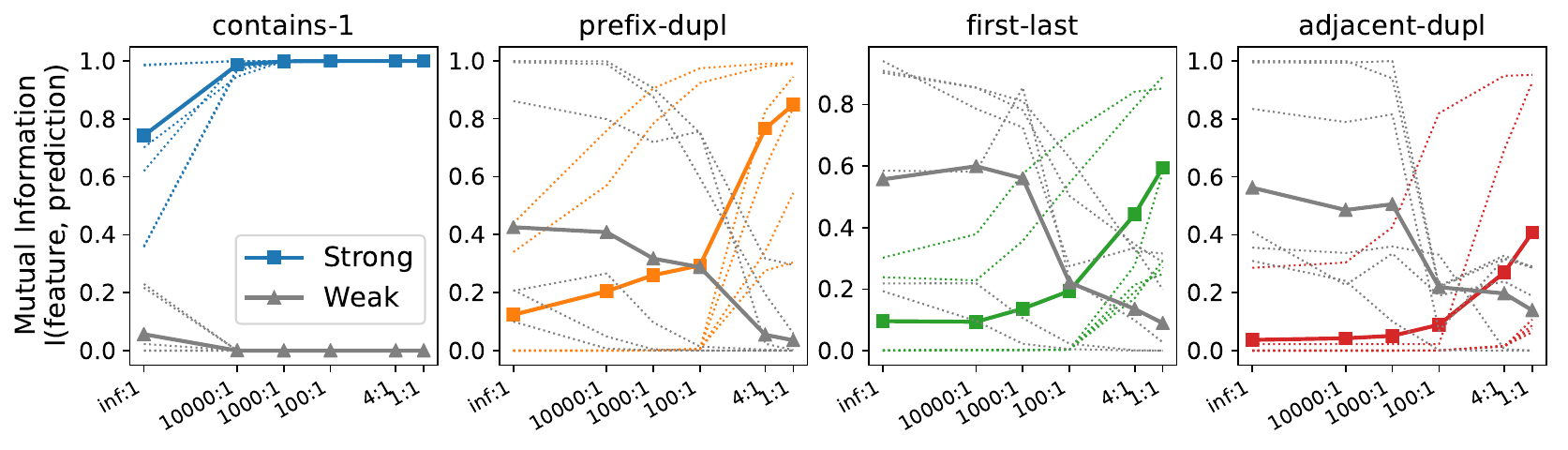}
                  \caption{Mutual information ($y$-axis) between presence of feature and model's prediction, as a function of dataset skew ($x$-axis). Skew $m:n$ means that the features occur together $m$ times for every $n$ times the weak feature occurs alone. Ideally, $\mathcal{I}(weak, pred)$ would be near $0$ and $\mathcal{I}(strong, pred)$ near $1$. Dark lines are aggregated over multiple simulations (in which we alter the frequency of \textit{strong-only} as discussed in \S\ref{sec:counterexamples}). %in the same way as we alter the number of \textit{weak-only} examples along the $x$-axis).
                  Deaggregated runs shown as thin lines; when there is variation, runs achieving high $\mathcal{I}(strong, pred)$ and low $\mathcal{I}(weak, pred)$ are those with higher numbers of \textit{strong-only} examples (discussed more in \S\ref{sec:skew}).}
    \label{fig:mi}
\end{figure*}

%\begin{figure*}[ht!]
%    \centering
%            \begin{subfigure}{\textwidth} 
%        \includegraphics[width=\linewidth]{figures/heatmap_weak}
%        \subcaption{Model's sensitivity to the weak feature, $I(label, weak\_feature)$}
%                   \label{fig:miweak}
%       \end{subfigure} \\
%        \begin{subfigure}{\textwidth} 
%        \includegraphics[width=\linewidth]{figures/heatmap_strong}
%        \subcaption{Model's sensitivity to the strong feature, $I(label, strong\_feature)$}
%                    \label{fig:mistrong}
%        \end{subfigure} 
%            \caption{Mutual information ($I$) between feature presence and model prediction \rohan{for the test data}. $I$ ranges from $0$ (low) to $1$ (high), but note that colors are flipped so that blue is always the desired direction (i.e. blue = lower for $I(prediction, weak)$ and blue = higher for $I(prediction, strong)$). Leftmost heatmaps show $I$ according to the training data.\ellie{Clean up a bit more.}}
%    \label{fig:mi}
%\end{figure*}

%We also note that
%, once controlling for the effects of specific counter example types,
%the effect of total training size is not significant, and that the effect of label skew ($P(label=1)$) impacts both metrics similarly. %\rohan{TODO: }Mutual information between the weak feature and the label in the training data has a significant negative effect on \textit{weak-only} accuracy but no significant effect on \textit{strong-only} accuracy.

\section{In-Depth Analyses}
\label{sec:analysis}
Regression analysis
%highlights several interesting trends, but has limitations and
cannot tell the complete story.
%of how data augmentation affects a model's decisions.
We thus take a closer look at the effect of hardness (\S\ref{sec:hardness}) and at the sometimes-negative effects associated with data augmentation (\S\ref{sec:heuristics} and \S\ref{sec:skew}). 

\subsection{Effect of Strong Feature's Hardness}
\label{sec:hardness}

In Section \ref{sec:regression}, we observed a relationship between the hardness of the strong feature and both the overall accuracy of the model and the effectiveness of counterexamples. To better understand this relationship, we look at the mutual information $\mathcal{I}$ between the model's prediction at test time and the presence of each feature. In a perfect model (i.e., one basing its decisions entirely on the strong feature), the presence of the weak feature would have no effect on the prediction, so $\mathcal{I}(weak, pred)\approx0$ and $\mathcal{I}(strong, pred)\approx1$. In Figure \ref{fig:mi}, for each feature, we look at $\mathcal{I}$ as a function of the relative frequency of \textit{weak-only} vs.\ \textit{strong-only} examples in the training data. Specifically, we plot $\mathcal{I}(weak, pred)$ and $\mathcal{I}(strong, pred)$ as a function of the ratio $m:n$ where $m$ is the number of \textit{both} examples and $n$ is the number of \textit{weak-only} examples. We also vary the number of \textit{strong-only} examples (\S\ref{sec:counterexamples}), and Figure \ref{fig:mi} shows the aggregated trends (across all numbers of \textit{strong-only} examples) as well as the deaggregated runs.

%Figure \ref{fig:mi} shows how $I$ varies across all combinations of numbers of \textit{weak-only} and \textit{strong-only} counterexamples. Note that, according to the training data (leftmost heat map in each row of Fig.\ \ref{fig:mi}) $I(label, strong)$ is always close to 1, and $I(label, weak)$ strictly decreases as we add counterexamples.  

There is a clear association between the hardness of the strong feature and the sensitivity of the model to the dataset distribution (the $m:n$ ratio). For example, for \texttt{contains-1}, even when the training data is pathologically skewed ($10^4$ \textit{both} examples for every $1$ \textit{weak-only} example), the model learns the desired associations: the strong feature alone determines the prediction. In contrast, for \texttt{adjacent-dupl}, even when the data is perfectly balanced with an equal number of \textit{both} and \textit{weak-only} examples, the model still does not exhibit the desired behavior. Rather, the model appears to unlearn the heuristic based on the weak feature ($I(weak, pred)\approx0$) but does not use the \textit{strong} feature as it should ($I(strong, pred) << 1$). Note, in some settings--when the number of \textit{strong-only} examples in the training data is very high--the model appears to learn the correct associations. We note that, in practice, this is likely to be a property of the training data/task that is outside of our control as practitioners (\S\ref{sec:counterexamples}).

\begin{figure*}[ht!]
    \centering
        \includegraphics[width=\linewidth]{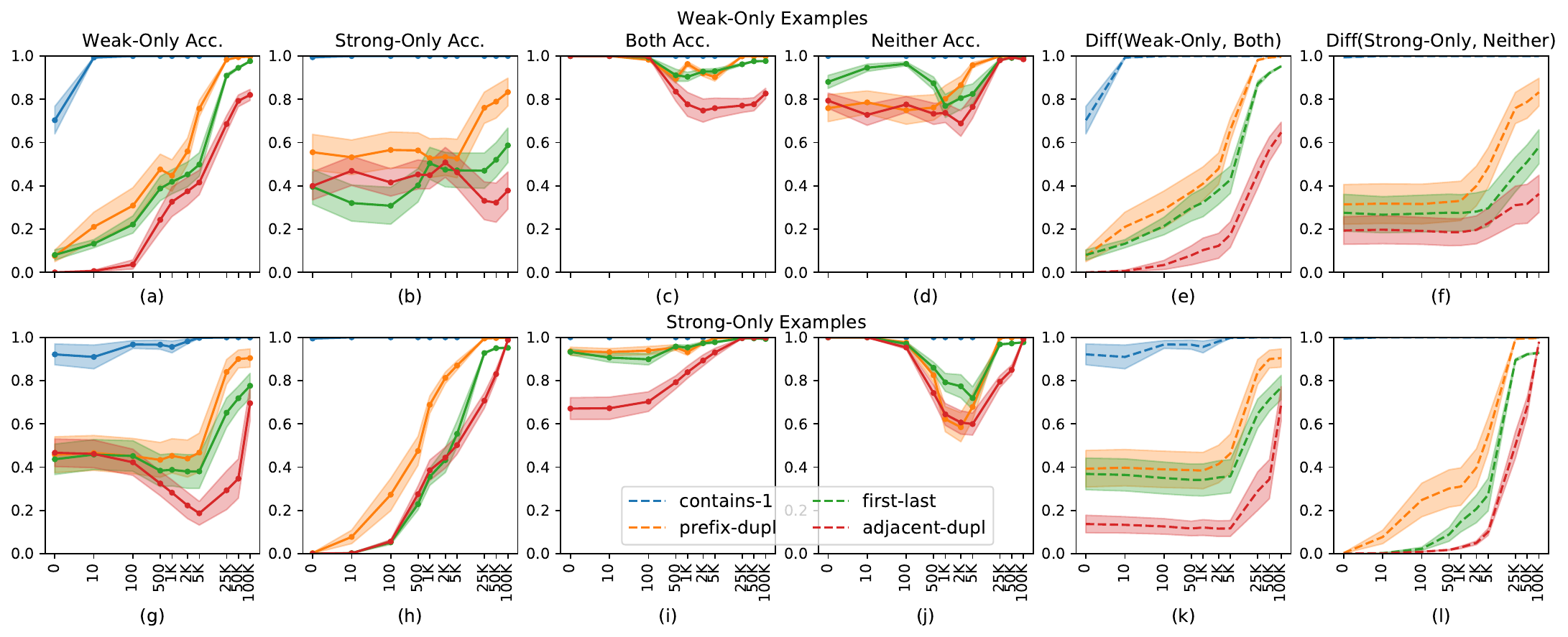}
            \caption{{\bf Left (a--d and g--j):} Accuracies of models for a given number of counterexamples of one type, averaged over all values of counterexamples of the other type. Graphs should be compared in terms of trajectories (increasing vs.\ decreasing) rather than absolute numbers along the $y$-axis ($0$ vs.\ $0.5$). This is because starting points on the $y$-axis are not always comparable. E.g., the difference in the starting points of (a) vs.\ (b) reflects that runs with high numbers of \textit{strong-only} examples favor \textit{strong-only} acc (b) more than they favor \textit{weak-only} acc (a). {\bf Right (e--f and k--l):} Differences between classification rates. See text for details.
            % I don't get anything from "These offset one another when only the weak feature is used".
            }
    \label{fig:line-plots}
\end{figure*}

\subsection{Nonlinear Trends}
\label{sec:heuristics}
Our regression analysis suggests that the effect of adding counterexamples is sometimes negative. To further investigate how accuracy changes as a function of the number and type of counterexample, we focus on the effect of adding one type of counterexample (\textit{weak-only} or \textit{strong-only}), averaging over all levels of the other type of counterexample (\textit{strong-only} or \textit{weak-only}, respectively). %Here, we investigate the trends across training configurations, and
In Section \ref{sec:skew}, we look at the deaggregated trends.

% \charliereplace{\paragraph{General Observations.} Figure \ref{fig:line-plots} shows aggregated trends for each of the different choices of strong property and for each accuracy metric. We clearly see the expected positive linear relationship between \textit{weak-only} examples and \textit{weak-only} accuracy, and between \textit{strong-only} examples and \textit{strong-only} accuracy, but the other relationships are more nuanced. For example, we see a distinct U-shaped curve when looking at the effect of \textit{strong-only} examples on \textit{weak-only} acc.(Fig. \ref{fig:line-plots}(g)), and when looking at the effect of either type of example on \textit{neither} acc. (Figs. \ref{fig:line-plots}(d,j)).}
% > two lines shorter

\paragraph{General Observations.} Figure \ref{fig:line-plots} shows aggregated trends over each feature and accuracy metric. We see the expected positive linear relationship between \textit{weak-only} examples and \textit{weak-only} accuracy, and between \textit{strong-only} examples and \textit{strong-only} accuracy. The other relationships are more nuanced: e.g., the effect of \textit{strong-only} examples on \textit{weak-only} accuracy (Fig.\ \ref{fig:line-plots}(g)), and the effect of either example type on \textit{neither} accuracy (Figs.\ \ref{fig:line-plots}(d,j)), appear to produce U-shaped curves. %There is some evidence of this U-curve as well when looking at the effect of \textit{weak-only} examples on \textit{strong-only} accuracy (Fig. \ref{fig:line-plots}(b)), though it is much less pronounced and varies substantially across properties.}{}

\paragraph{Relationship to Decision Rules.} 
To understand these trends better, we attempt to connect them to the ``rules'' a model could use. For example, if a model uses the heuristic $weak \to 1$ and $\neg weak \to 0$, we'd expect low \textit{strong-only} and \textit{weak-only} accuracy and high \textit{both} and \textit{neither} accuracy. Table \ref{tab:heuristics} summarizes the rules the model might use if it represents the weak feature $w$ or the strong feature $s$. This discussion is intended to be qualitative and informal, to summarize our interpretation of the quantitiative trends in Fig.\ \ref{fig:line-plots}.

\begin{table*}[ht!]
\centering
\small
\begin{tabular}{cr|cccc|cccc|cc}
\toprule
& & \multicolumn{4}{c|}{Accuracy Metrics} & \multicolumn{6}{c}{Possible Rules}\\
%\midrule
\cmidrule(lr){3-6}\cmidrule(lr){7-12}
&			&  wk-only 		& both 		& str-only 		& neither 
%			& \rotatebox[origin=c]{60}{$w\to1$}
%			& \rotatebox[origin=c]{60}{$\neg w\to 0$}
%			& \rotatebox[origin=c]{60}{$w\to0$}
%			& \rotatebox[origin=c]{60}{$\neg w\to 1$}
%			& \rotatebox[origin=c]{60}{$s\to 1$}
%			& \rotatebox[origin=c]{60}{$\neg s\to 0$}\\
			& $w:1$
			& $\neg w:0$
			&$w:0$
			& $\neg w:1$
			& $s:1$
			& $\neg s:0$\\
%			& ${w, \neg s}$	 	& ${w, s}$ 	& ${\neg w, s}$ 		& ${\neg w, \neg s}$
%\cmidrule(lr){3-6}\cmidrule(lr){7-12}
\midrule
\multirow{6}{*}{\rotatebox[origin=c]{90}{Possible Rules}} & $w\to1$		& 	\cellcolor{red!25}\xmark	&\cellcolor{green!25}\cmark	& 	--					& --			
 &\cellcolor{gray!50}&&&&&\\
& $\neg w\to0$	& 	--					& 	--					& \cellcolor{red!25}\xmark		& \cellcolor{green!25}\cmark		
 & &\cellcolor{gray!50}&&&&\\
&  $w\to0$		& 	\cellcolor{green!25}\cmark	&  \cellcolor{red!25}\xmark		& 	--					& --			
 & &&\cellcolor{gray!50}&&&\\
&  $\neg w\to1$	& 	--					& 	--					& \cellcolor{green!25}\cmark	& \cellcolor{red!25}\xmark		
 & &&&\cellcolor{gray!50}&&\\
 & $s\to1$		& 	--					& \cellcolor{green!25}\cmark	& \cellcolor{green!25}\cmark	& --			
 & &&&&\cellcolor{gray!50}&\\
 & $\neg s\to0$ 	& \cellcolor{green!25}\cmark	& --						& --						& \cellcolor{green!25}\cmark			
 & &&&&&\cellcolor{gray!50}\\
\midrule
 &\multicolumn{11}{c}{Adding Weak-Only Examples} \\
\midrule
\multirow{4}{*}{\rotatebox[origin=c]{90}{Data Aug.}} & None 	& 	\cellcolor{red!25}\xmark	&	\cellcolor{green!25}\cmark	&	\cellcolor{red!25}\xmark	& \cellcolor{green!25}\cmark 
& \cellcolor{gray!50}&\cellcolor{gray!50}&&&&\\
& Low  	& 	\cellcolor{green!25}$\nearrow$	&	\cellcolor{red!25}$\searrow/-$	&\cellcolor{green!25}$\nearrow/-$	& \cellcolor{red!25}$\searrow/-$
&&\cellcolor{gray!25}&\cellcolor{gray!25}&\cellcolor{gray!25}&&\\
& Mid    & 	\cellcolor{green!25}$\nearrow$	&	\cellcolor{green!25}$\nearrow$	&\cellcolor{red!25}$\searrow/-$	& \cellcolor{green!25}$\nearrow$
&&\cellcolor{gray!25}&&&\cellcolor{gray!25}&\cellcolor{gray!25}\\
& High 	& 	\cellcolor{green!25}\cmark		&	\cellcolor{green!25}\cmark	&\cellcolor{red!25}\xmark/$\nearrow$ & \cellcolor{green!25}\cmark
&&\cellcolor{gray!25}&&&\cellcolor{gray!25}&\cellcolor{gray!50}\\
\midrule
& \multicolumn{11}{c}{Adding Strong-Only Examples} \\
\midrule
\multirow{4}{*}{\rotatebox[origin=c]{90}{Data Aug.}} & None  	& 	\cellcolor{red!25}\xmark	&	\cellcolor{green!25}\cmark	&	\cellcolor{red!25}\xmark	& \cellcolor{green!25}\cmark 
& \cellcolor{gray!50}&\cellcolor{gray!50}&&&&\\
& Low  	& \cellcolor{red!25}$\searrow/-$	&	\cellcolor{green!25}$\nearrow$		&\cellcolor{green!25}$\nearrow$ &	\cellcolor{red!25}$\searrow$
&\cellcolor{gray!25}&&&\cellcolor{gray!25}&&\\
& Mid  	& 	\cellcolor{green!25}$\nearrow$	&	\cellcolor{green!25}$\nearrow$	&\cellcolor{green!25}$\nearrow$	& \cellcolor{green!25}$\nearrow$
&&&&&\cellcolor{gray!25}&\cellcolor{gray!25}\\
& High 			& 	\cellcolor{green!25}\cmark	&	\cellcolor{green!25}\cmark	&\cellcolor{green!25}\cmark	& \cellcolor{green!25}\cmark
&&&&&\cellcolor{gray!50}&\cellcolor{gray!50}\\
\bottomrule
\end{tabular}
\caption{Summary of rule-like behavior of model at different levels of data augmentation. Symbols (\cmark, \xmark) denote stable points, at the beginning or when the model seems to have converged. Arrows ($\nearrow$, $\searrow$) denote trends, when the model is shifting behavior. Slashes indicate ambiguity, when different features behave notably differently. Dark gray means that the model is behaving consistently with a given rule. Lighter gray means error patterns are partially consistent with and/or trending toward the rule's error patterns.}% \rohan{Okay! I think I'm on board in general.} \rohan{Quibbles for adding weak: strong starting at wrong. Quibbles for adding strong: weak starting at wrong, the middle cells for both could maybe be $\nearrow/-$.}}
\label{tab:heuristics}
\end{table*}

\subparagraph{Adding \textit{strong-only} examples.} We look first at the effects of adding \textit{strong-only examples}. Before adding any counterexamples, the model achieves low  \textit{strong-only} and  \textit{weak-only} accuracies, and high \textit{both} and \textit{neither} accuracies. This behavior is consistent with the rule $w\to 1, \neg w \to 0$. Once we begin adding counterexamples, we see that learning can be divided into two clear phases. First, at low levels of data augmentation (up to 2K examples, around $1\%$ of the training data), \textit{weak-only} and \textit{neither} accuracies are either flat or decreasing while \textit{strong-only} and \textit{both} accuracies are increasing. This behavior would be consistent with a shift from using $\neg w \to 0$ to using $\neg w \to 1$. Second, at higher levels of augmentation, the trajectory shifts, and all accuracy metrics begin increasing in tandem. This behavior is not consistent with any combination of rules that rely only on $w$, and thus suggests that the model is beginning to represent and to use the strong feature. This story is summarized in Table \ref{tab:heuristics}.

Figures \ref{fig:line-plots}(k,l) tell this same story a bit differently, by looking at the (absolute) difference between $\hat{P}({1}|\textit{weak-only})$ and $\hat{P}({1}|\textit{both})$ and at the difference between $\hat{P}({1}|\textit{strong-only})$ and $\hat{P}({1}|\textit{neither})$. Here, $\hat{P}({1})$ 
is the rate at which the model predicts $1$. When the model is only using the weak feature in its decisions, these metrics necessarily move against each other: e.g., it is not possible to differentiate between \textit{both} and \textit{weak-only} unless the strong feature is somehow represented. Thus, during the first phase (low levels of data augmentation), we see that these differences are flat, suggesting that changes in these metrics exactly offset one another and thus the role of the strong feature in the models' predictions is not changing. At higher levels of data augmentation, these metrics stop competing, and we see the differences going to one, suggesting the model cannot be relying on the weak feature alone.

\begin{figure*}[ht!]
    \centering
         \includegraphics[width=\linewidth]{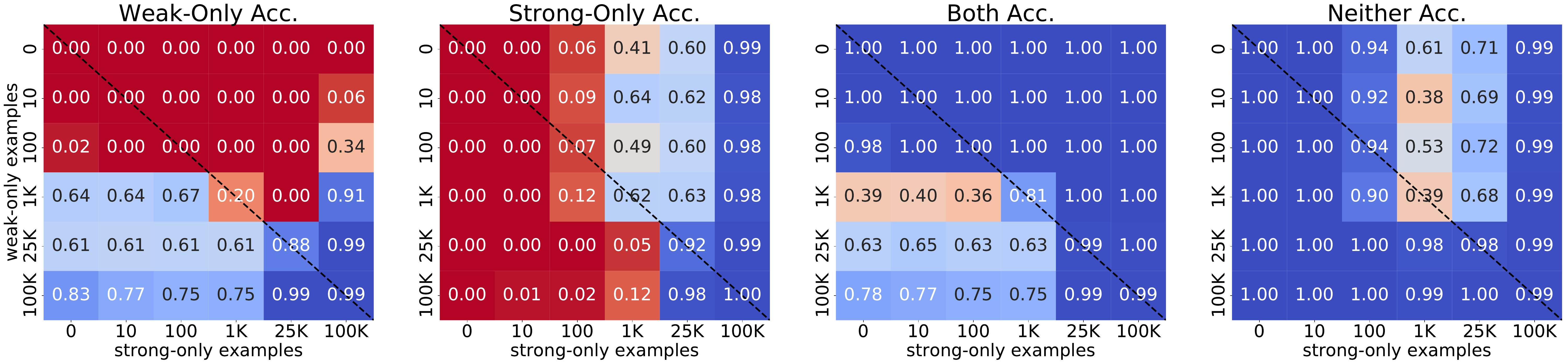}
       \caption{Accuracies for \texttt{adjacent-dupl} with \# \textit{weak-only} $\times$ \# \textit{strong-only} examples.}
    \label{fig:accuracy-heatmaps}
\end{figure*}

\subparagraph{Adding \textit{weak-only} examples.} The trends  for \textit{weak-only} examples (Figs.\ \ref{fig:line-plots}(a--f)) are similar to those for \textit{strong-only} examples. However, curves show higher variance in general, and when the model shifts from the first phase (where it moves from one weak-feature-based rule to another) to the second (where it begins using the strong feature) the \textit{strong-only} accuracy learning curve is much flatter. This is exacerbated when the strong feature is harder to extract (i.e., \texttt{first-last} and \texttt{adjacent-dupl}). As a result, at the levels of data augmentation we explore, we never see the model achieve high \textit{strong-only} accuracy on these features. However, we do see the model reach perfect performance on all other accuracy metrics. This is consistent with the model learning to use the strong feature in some cases, but not yet abandoning the $\neg w \to 0$ heuristic.

\subsection{Interaction Between Example Types}
\label{sec:skew}

Above, we focused on the effect of adding one type of counterexample, largely ignoring how the number counterexamples of other type already present in the training data impact results. The two counterexample types are likely to complement one another, and different learning problems warrant different assumptions about how the initial training set is distributed. Thus, it's important to understand how the effectiveness of data augmentation changes as we vary assumptions about the initial distribution of the training data. 

Figure \ref{fig:accuracy-heatmaps} shows each accuracy metric across all combinations of \# \textit{strong-only} examples $\times$ \# \textit{weak-only} examples for the \texttt{adjacent-dupl} feature.\footnote{Other features given in Supplementary Material (\S \ref{sec:acc}).} We see that when the number of \textit{strong-only} examples is extremely low ($0$ or $10$), no number of \textit{weak-only} examples affects \textit{strong-only} error, and vice-versa when then number of \textit{weak-only} examples is extremely low. Perhaps more interestingly, we see that when the number of \textit{strong-only} examples is low but not negligible (around $1\%$ of training data), adding \textit{weak-only} examples can hurt rather than help. We hypothesize that around this level of data skew, the model learns something other than just the weak feature but has insufficient data to learn the strong feature. Thus, it uses features that are stronger during training but fail to generalize to test (e.g. character bigrams).

\section{Discussion and Future Work}

Our analysis reveals a few trends that warrant further investigation, as they may have implications for the effectiveness of data augmentation and the behavior of models trained via these methods. First, the U-shaped learning curves observed suggest that low levels of data augmentation are likely only to improve some metrics  at the expense of others. There are two notable properties of the learning problem that appear to make this problem more pronounced: a) when only \textit{weak-only} examples are added and b) when the strong feature is significantly more difficult to extract from the input than the targeted weak feature. These two properties are significant as they correspond to the typical setting in which we'd expect to use data augmentation in practice. One straightforward takeaway from this is the importance of thoroughly evaluating models trained via data augmentation. Evaluating only on the targeted phenomenon (\textit{weak-only} error) is likely to over-state the gains; evaluating only on aggregated metrics (accuracy on standard test sets) is likely to hide interesting win-loss patterns.

Second, the relationship between the strong feature's ``hardness'' and the model's sensitivity to counterexamples (of both types) suggests promising directions for future work. In particular, our experiments suggested that current data augmentation practices--i.e. adding small numbers of \textit{weak-only} examples--can be effective as long as the strong feature is not too much harder than the weak feature to extract from the input. This suggests that further investing in representation learning, and specifically prioritizing the \textit{extractability} of a feature in the input representation, could yield better outcomes than investing in data augmentation alone. Such a direction could be interesting from both a practical and a theoretical perspective. 

Finally, there are limitations in our setup. We focus on a simple model and task, and it remains to be seen whether similar patterns hold for e.g., large pretrained language models. Also, we use a fixed training size ($200$K) and cannot say whether the trends we observed across ``low'' vs.\ ``high'' levels of data augmentation are better understood as absolute numbers (low=$2$K counterexamples) or as a proportion of training data (low=$1\%$) (or rather both). Finally, we focused only on the case in which there is a single strong feature and a single weak feature in the data. In real language tasks, this assumption would never hold. 
\section{Related Work}
\paragraph{Data Augmentation.}
A wave of recent work has constructed evaluation sets composed of ``adversarial examples'' (or ``challenge examples'' or ``probing sets'') to analyze weaknesses in the decision procedures of neural NLP models \citep[and others]{jia2017adversarial,glockner2018,dasgupta2018evaluating,gururangan2018annotation,poliak2018hypothesis}. Our work is motivated by the subsequent research that asks if adding such examples to training data help improve robustness.
\citet{liu-etal-2019-inoculation} show that fine-tuning on small challenge sets can sometimes (but not always) help model performance. Similar approaches have been explored for improving syntactic parsing \citep{elkahky-etal-2018-challenge}, improving NLI models' handling of syntactic \citep{min2020augmentation} and semantic \citep{poliak-etal-2018-collecting} phenomena, and mitigating gender biases \citep{zmigrod-etal-2019-counterfactual,zhao-etal-2018-gender,zhao2019gender,hall-maudslay-etal-2019-name,lu2018gender}. \citet{nie-etal-2020-adversarial} and \citet{Kaushik2020Learning} also explore augmenting training sets with a human-in-the-loop.

%\subsection{Adversarial Robustness}

\paragraph{Feature Representations and Robustness.}

A related body of recent work asks which features are extracted by neural language models, in particular whether SOTA models represent ``deeper'' syntactic and semantic features. Work in this vein has shown that pretrained language models encode knowledge of syntax, using a range of techniques including supervised ``diagnostic classifiers'' \citep{tenney2018what,conneau-etal-2018-cram,hewitt-manning-2019-structural}, classification performance on targeted stimuli \citep{linzen-etal-2016-assessing,goldberg2019assessing}, attention maps/visualizations \citep{voita-etal-2019-analyzing,serrano-smith-2019-attention}, and relational similarity analyses \citep{chrupala-alishahi-2019-correlating}. \citet{geiger-etal-2019-posing} attempts to quantify how much data a model should need to learn a given deeper feature. We contribute to this literature by asking under what conditions we might expect deeper features to be extracted and used.  
%, focusing in particular on the role that the training distribution plays in encouraging models to learn deeper structure. Related in spirit to our toy data approach is recent work which attempts to quantify how much data a model should need to learn a given deeper feature \citep{geiger-etal-2019-posing}. Still other related work explores ways for encouraging models to learn structure which do not rely on data augmentation, e.g. by encoding inductive biases into model architectures \citep{bowman-etal-2015-recursive,andreas2016neural} in order to make ``deep'' features more readily extractable, or by designing training objectives that incentivize the extraction of specific features \citep{swayamdipta2017frame,niehues-cho-2017-exploiting}. Exploring the effects these modeling changes on the results presented in this paper is an exciting future direction.

Adversarial robustness examines how small perturbations in inputs can cause models to make wrong predictions \citep{ribeiro2018semantically, iyyer2018adversarial, hsieh2019robustness, jia2019certified}, or to change their output \citep{alzantot2018generating} or internal representations \citep{hsieh2019robustness}. In NLP, such perturbations often involve word-level \citep{alzantot2018generating, hsieh2019robustness, jia2019certified} or sentence-level \citep{ribeiro2018semantically, iyyer2018adversarial} paraphrasing. \citet{ilyas2019adversarial} make a distinction between useful features (that generalize well) and those that are robustly-useful (that generalize well, even if an example is adversarially perturbed). \citet{madry2017towards,athalye2018obfuscated} investigate robustness by giving adversaries access to model gradients.

% Other approaches for understanding improving robustness involve varying assumptions about whether/to what extent adversaries have access to the model's gradients . %And related to this work by \citet{ilyas2019adversarial}, there's been significant recent interest in training models such that they're robust to adversarial examples and in building adversarial datasets that foil such defenses. Highlighting just two recent papers, \citet{madry2017towards} describe training that's robust against adversaries with access to a model's gradients, while \citet{athalye2018obfuscated} show that many defenses are ``obfuscating'' their gradients in a way that can be exploited. 

%\subsection{Encoding Structure in NLP Models}

\paragraph{Generalization of Neural Networks}
A still larger body of work studies feature representation and generalization in neural networks. \citet{mangalam2019deep} show that neural networks learn ``easy'' examples (as defined by shallow ML model performance) before they learn ``hard'' examples. \citet{DBLP:journals/corr/ZhangBHRV16} and \citet{Arpit:2017:CLM:3305381.3305406} show that neural networks with good generalization performance can memorize noise, suggesting that such models might have an inherent preference to learn more general features. Finally, there is ongoing theoretical work that characterizes the ability of over-parameterized networks to generalize in terms of complexity \citep{neyshabur2018the} and implicit regularization \citep{blanc2019implicit}.

\section{Conclusion}
We propose a framework for simulating the effects of data augmentation in NLP and use it to explore how training on counterexamples impacts model generalization. Our results suggest that adding counterexamples in order to encourage a model to ``unlearn'' weak features is likely to have the immediately desired effect (the model will perform better on examples that look similar to the generated counterexamples), but the model is unlikely to shift toward relying on stronger features in general. Specifically, in our experiments, the models trained on data augmented with a small number of counterexamples ($<100$K) still fail to correctly classify examples which contain only the strong feature. We see also that data augmentation may become less effective as the underlying strong features become more difficult to extract.

\bibliography{main}
\bibliographystyle{acl_natbib}

\clearpage
\appendix

\section{Ablations} \label{sec:ablations}
\begin{figure*}
    \centering
         \includegraphics[width=\linewidth]{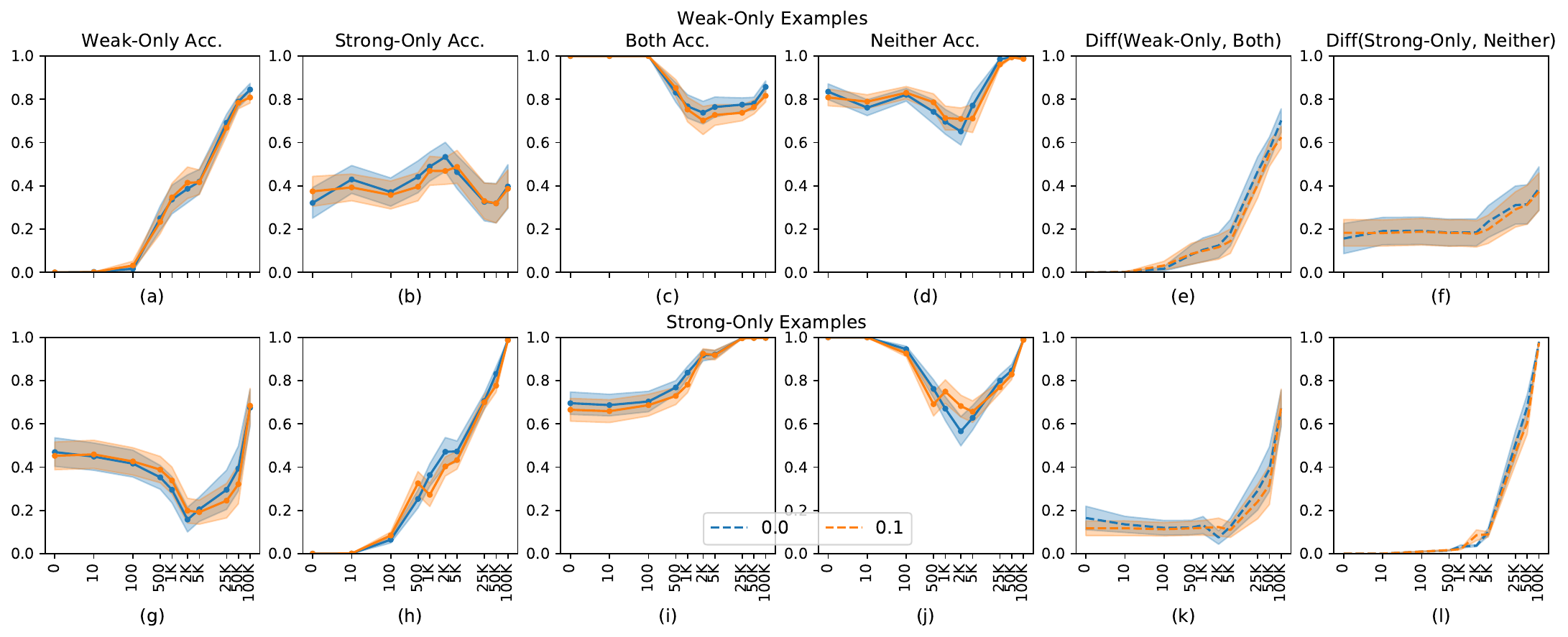}
       \caption{Dropout. We test dropout of 0 (original) and 0.1. Dropout is added after the first linear layer in the decoder.}
    \label{fig:dropout}
\end{figure*}

\begin{figure*}[h!]
    \centering
         \includegraphics[width=\linewidth]{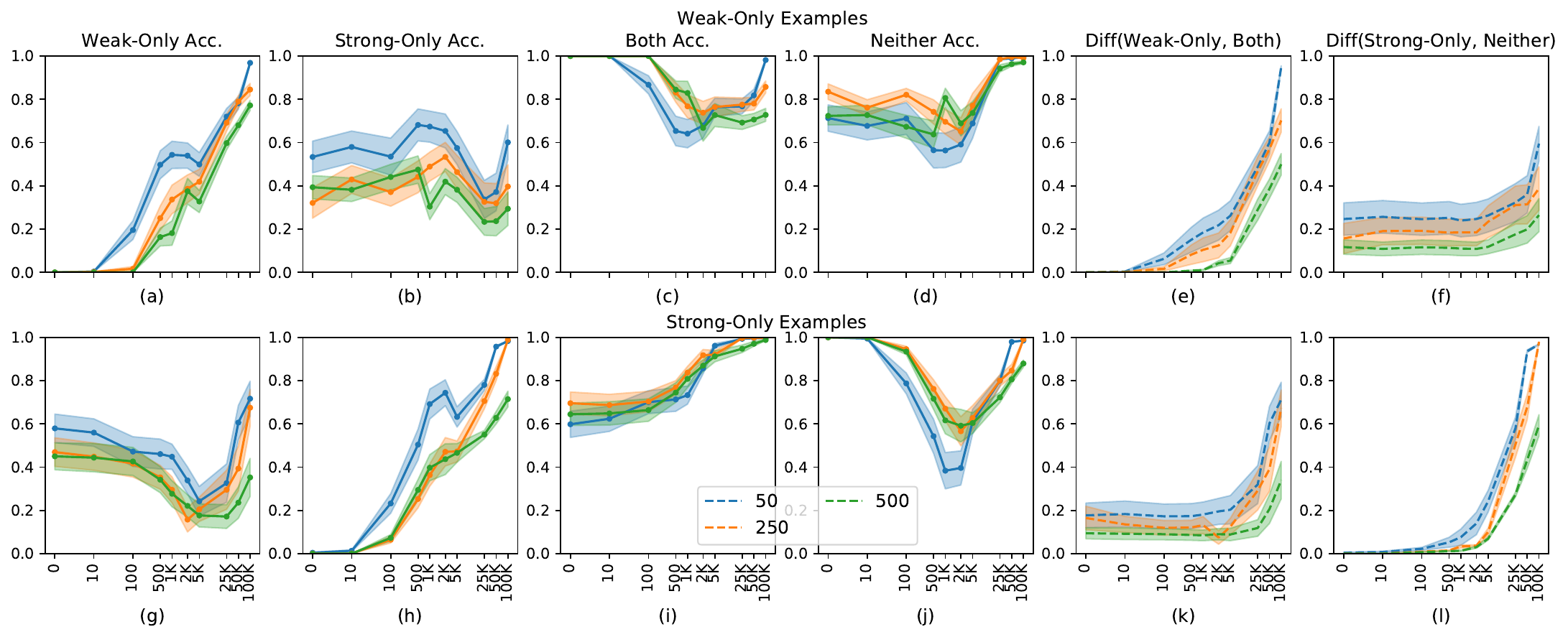}
       \caption{Embedding and hidden size. We test embedding and hidden sizes of 50, 250 (original), and 500.}
    \label{fig:hidden}
\end{figure*}

\begin{figure*}[h!]
    \centering
         \includegraphics[width=\linewidth]{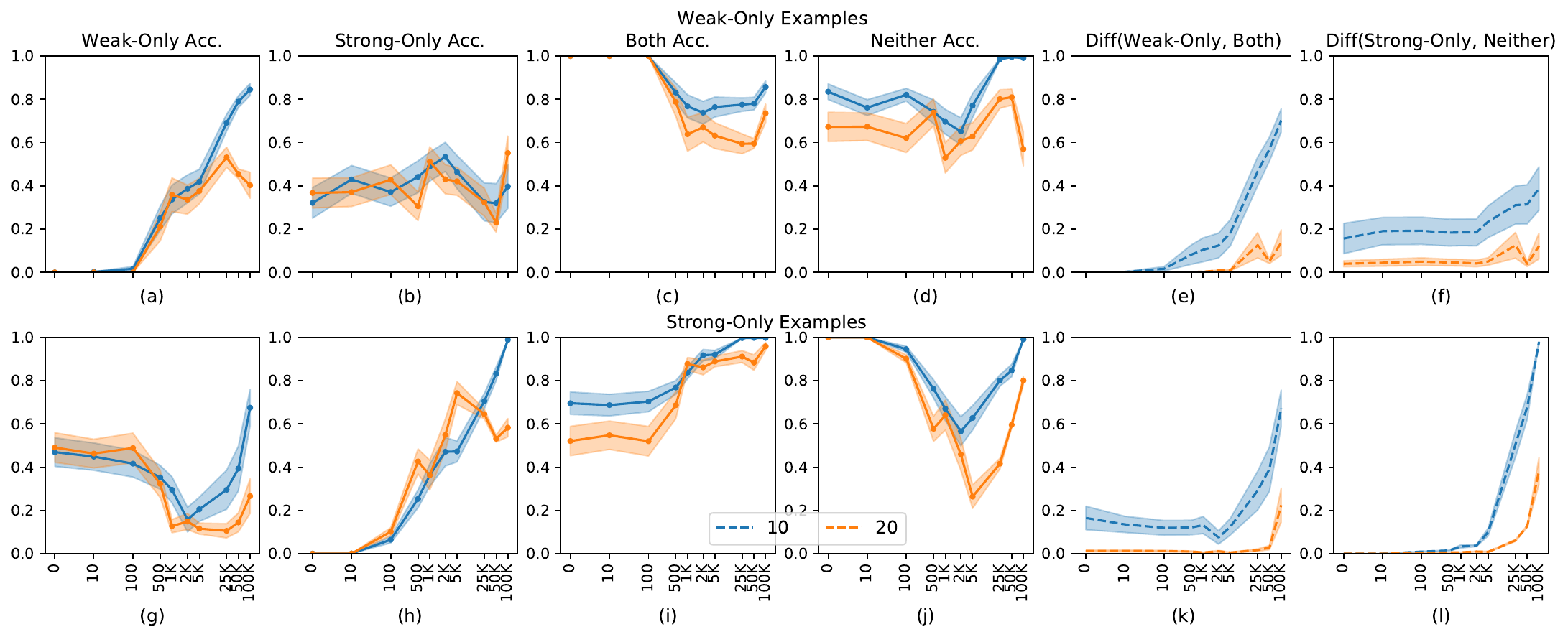}
       \caption{Sequence length. We test sequence lengths of 10 (original) and 20.}
    \label{fig:seq_length}
\end{figure*}

\begin{figure*}[h!]
    \centering
         \includegraphics[width=\linewidth]{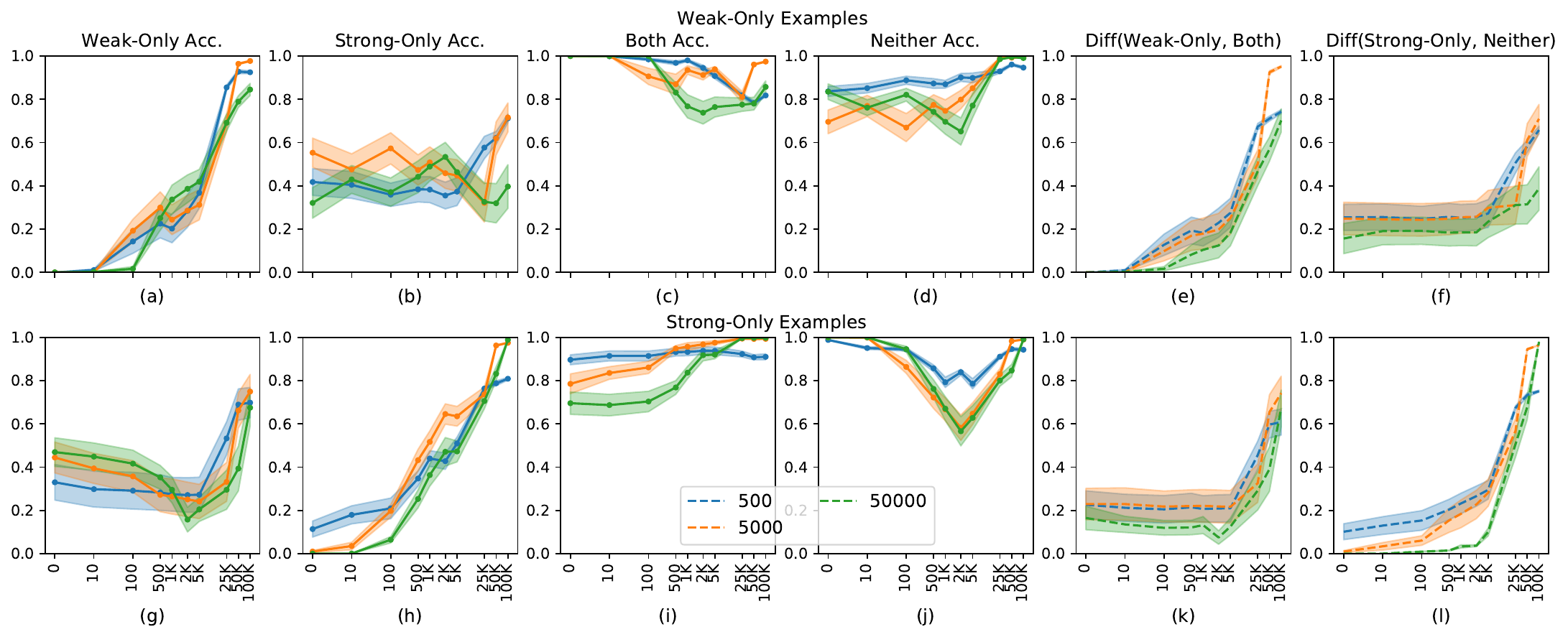}
       \caption{Vocab size. We test vocabulary sizes of 500, 5K, and 50K (original).}
    \label{fig:vocab}
\end{figure*}

We fix most system parameters in our experiments; Here we show that the trends hold under other parameters. We only use one random seed and one feature (\texttt{adjacent-dupl}). See Figures \ref{fig:dropout}-\ref{fig:vocab}. Like Figures \ref{fig:mi} and \ref{fig:line-plots}, accuracies are for a given number of counterexamples of one type, averaged over all values of counterexamples of the other type.

% \begin{figure*}[h!]
%     \centering
%              \includegraphics[width=\linewidth]{ablation_figures/big-figure-sample.pdf}
%       \caption{Sampling method. In all our experiments the symbols were sampled uniformly (besides those forming the given pattern). Here we change this modeling assumption and sample the symbols according to a zipfian distribution. We still keep the train/test symbols disjoint; we sample according to Zipf's Distribution over the symbols present in the given partition. Results are similar across the two settings; An exception is (j) where performance declines as the number of strong-only examples increases.}
%     \label{fig:sample}
% \end{figure*}


\section{De-Aggregated Results} \label{sec:de-agg}
\begin{figure*}[h!]
    \centering
         \includegraphics[width=\linewidth]{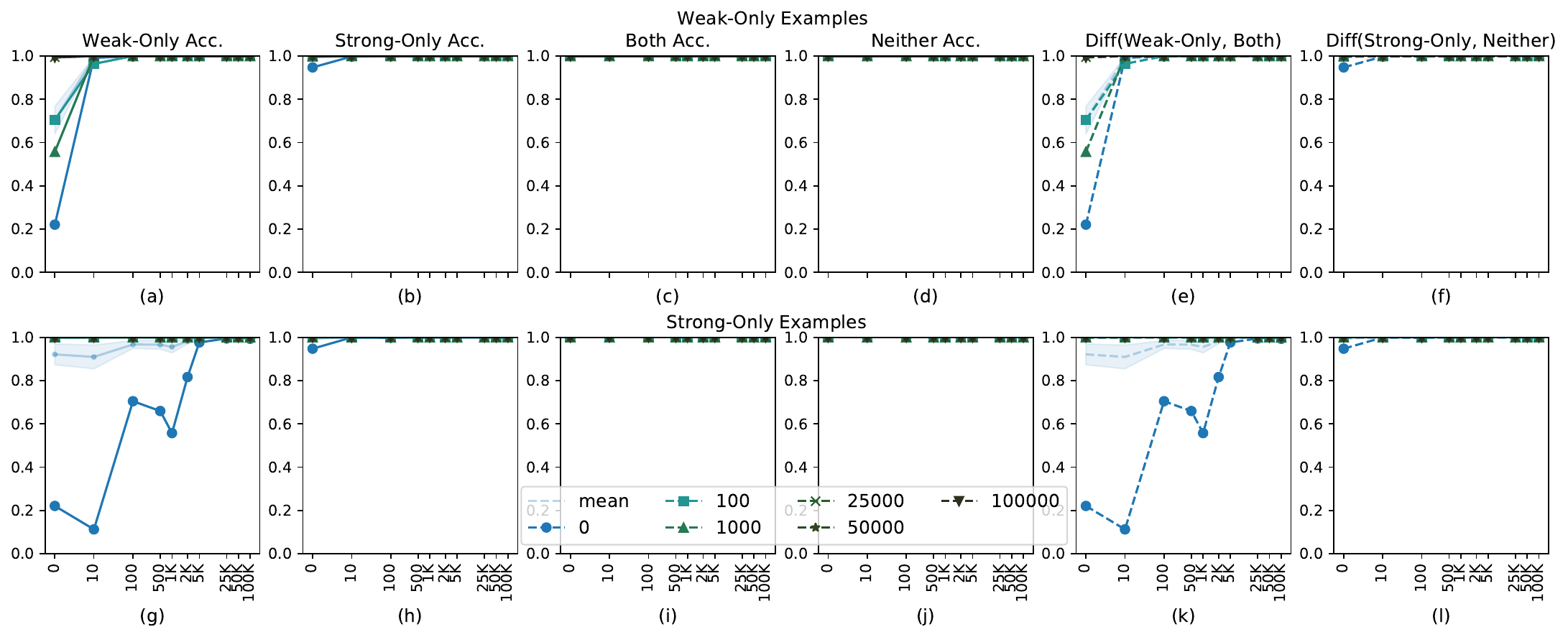}
       \caption{Curves for \texttt{contains-1} where the number \textit{weak-only} examples varies for different settings of \textit{strong-only} examples (top row) and vice versa (bottom row).}
    \label{fig:de1}
\end{figure*}

\begin{figure*}[h!]
    \centering
         \includegraphics[width=\linewidth]{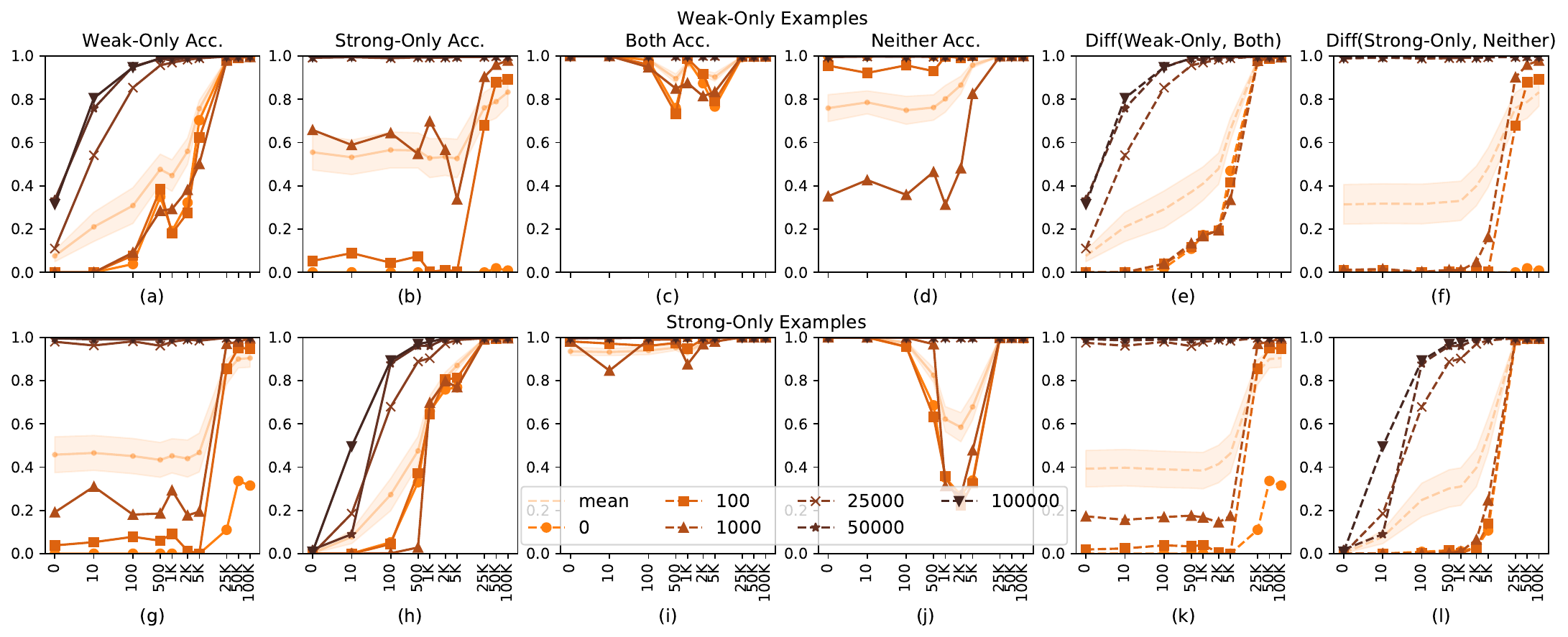}
       \caption{Curves for \texttt{prefix-dupl} where the number \textit{weak-only} examples varies for different settings of \textit{strong-only} examples (top row) and vice versa (bottom row).}
    \label{fig:de2}
\end{figure*}

\begin{figure*}[h!]
    \centering
         \includegraphics[width=\linewidth]{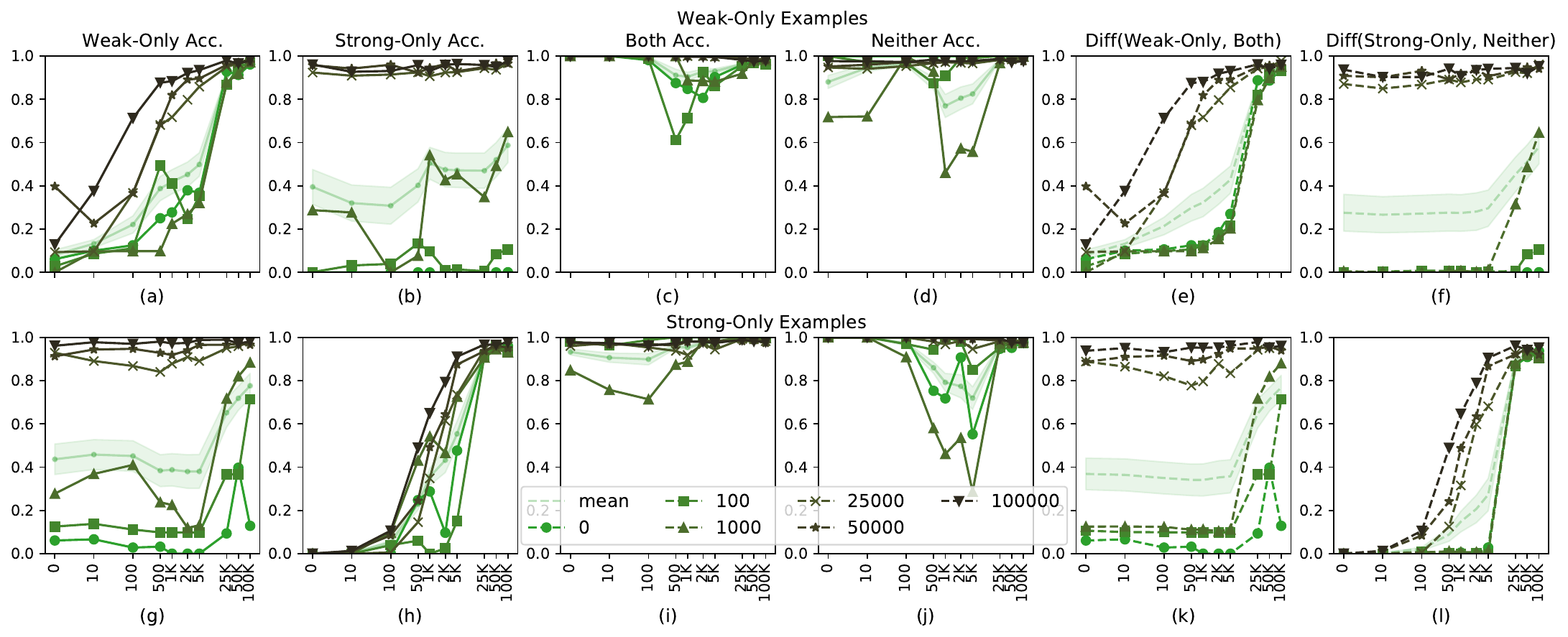}
       \caption{Curves for \texttt{first-last} where the number \textit{weak-only} examples varies for different settings of \textit{strong-only} examples (top row) and vice versa (bottom row).}
    \label{fig:de3}
\end{figure*}

\begin{figure*}[h!]
    \centering
         \includegraphics[width=\linewidth]{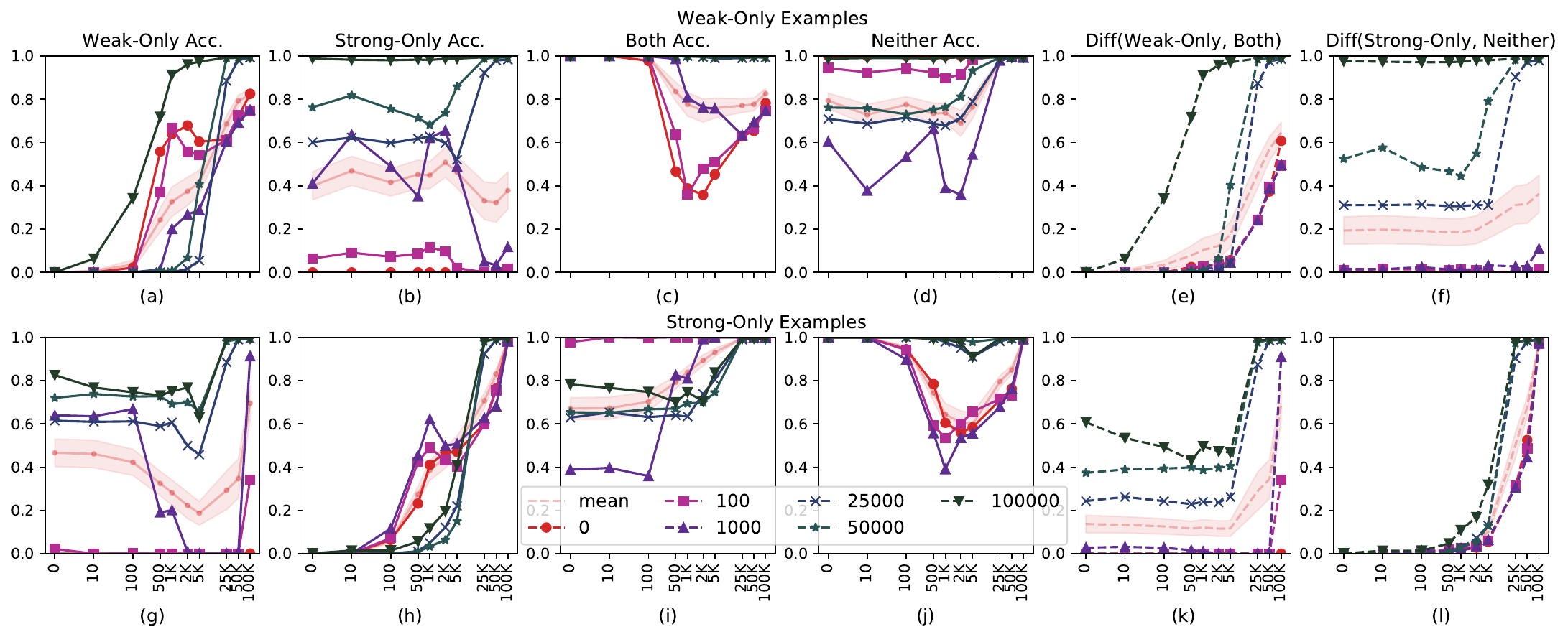}
       \caption{Curves for \texttt{adjacent-dupl} where the number \textit{weak-only} examples varies for different settings of \textit{strong-only} examples (top row) and vice versa (bottom row).}
    \label{fig:de4}
\end{figure*}

We de-aggregate the results over the number of counterexamples and inspect them for each feature. See Figures \ref{fig:de1}-\ref{fig:de4}.


\section{More Accuracies} \label{sec:acc}
\begin{figure*}[h!]
    \centering
         \includegraphics[width=\linewidth]{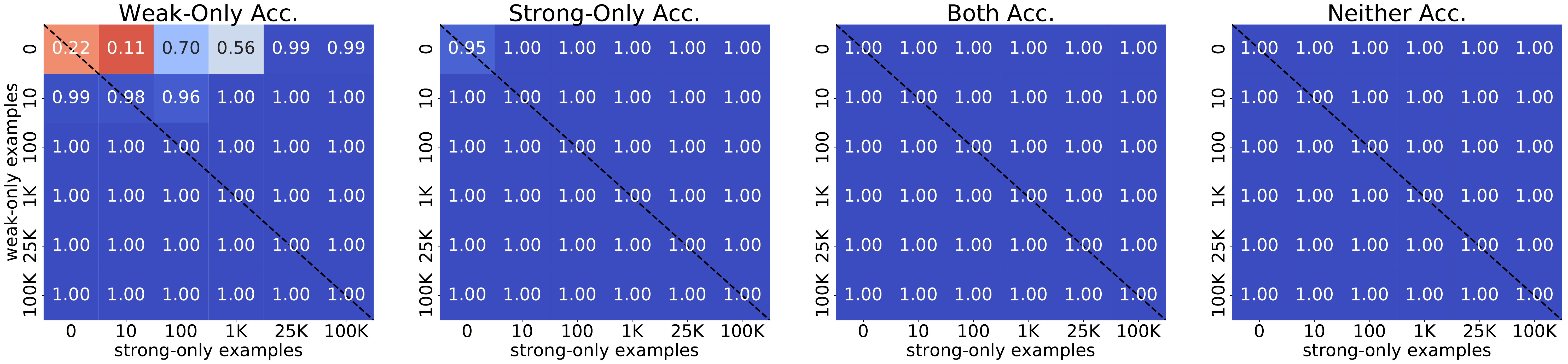}
      \caption{Accuracies for \texttt{contains-1} with \# \textit{weak-only} $\times$ \# \textit{strong-only} examples.}
    \label{fig:de5}
\end{figure*}

\begin{figure*}[h!]
    \centering
         \includegraphics[width=\linewidth]{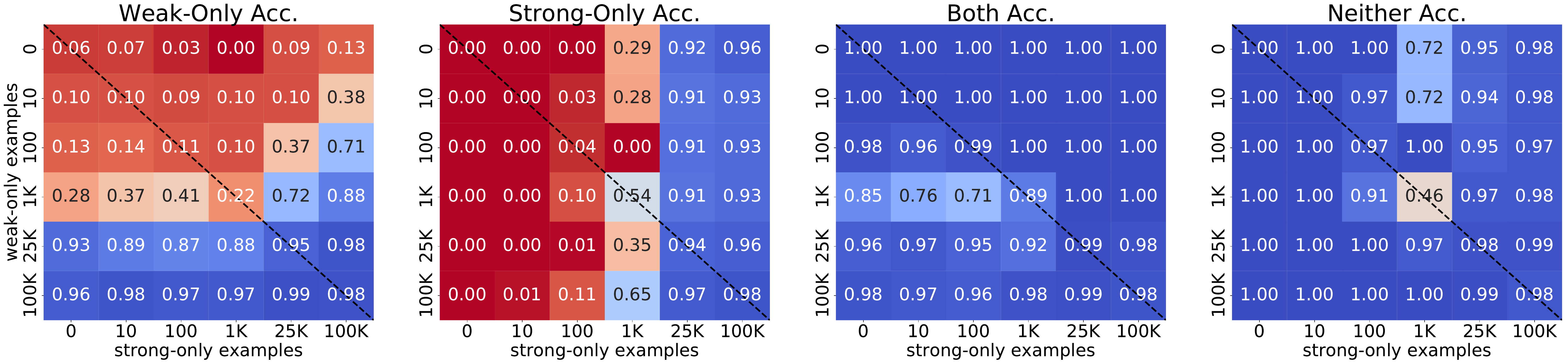}
      \caption{Accuracies for \texttt{first-last} with \# \textit{weak-only} $\times$ \# \textit{strong-only} examples.}
    \label{fig:de6}
\end{figure*}

\begin{figure*}[h!]
    \centering
         \includegraphics[width=\linewidth]{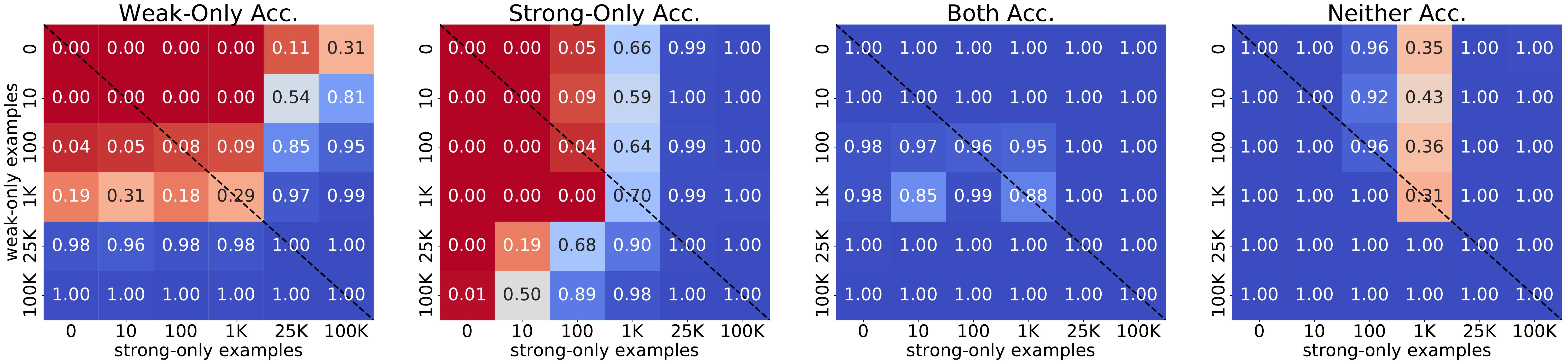}
      \caption{Accuracies for \texttt{prefix-dupl} with \# \textit{weak-only} $\times$ \# \textit{strong-only} examples.}
    \label{fig:de7}
\end{figure*}

Figures \ref{fig:de5}-\ref{fig:de7} show each accuracy metric across all combinations of \# \textit{strong-only} examples $\times$ \# \textit{weak-only} examples for the various features.


\section{More Regression Results} \label{sec:reg}
See the additional regression results in Tables \ref{tab:mains-interactions} and \ref{tab:mains-all-metrics}.

\begin{table*}[ht!]
\centering
\small
\setlength{\tabcolsep}{.5em}
\begin{tabular}{lcrlrl}
\toprule
&  & \multicolumn{2}{c}{\textit{weak-only} acc.} & \multicolumn{2}{c}{\textit{strong-only} acc.} \\
&  & \multicolumn{2}{c}{$R^2=0.82$}&\multicolumn{2}{c}{$R^2=0.84$} \\
\cmidrule(lr){3-4} \cmidrule(lr){5-6}
& $\sigma$ & \multicolumn{1}{c}{$\beta$} & \multicolumn{1}{c}{$p$} & \multicolumn{1}{c}{$\beta$} & \multicolumn{1}{c}{$p$} \\
%\cmidrule(lr){2-2} \cmidrule(lr){3-4} \cmidrule(lr){5-6}
\midrule
contains-1 &    &{$0.89$}&{$0.00$*}&{$0.90$}&{$0.00$*} \\ 
prefix-dupl &    &{$-0.07$}&{$0.01$*}&{$-0.01$}&{$0.60$} \\ 
first-last &    &{$-0.26$}&{$0.00$*}&{$-0.43$}&{$0.00$*} \\ 
adjacent-dupl &    &{$-0.58$}&{$0.00$*}&{$-0.48$}&{$0.00$*} \\ 
\midrule
log strong-only ex. & $1.5$&{$-0.22$}&{$0.00$*}&{$-0.15$}&{$0.00$*} \\ 
\hspace{2mm}pfx-dup &    &{$0.28$}&{$0.00$*}&{$0.86$}&{$0.00$*} \\ 
\hspace{2mm}first-last &    &{$0.15$}&{$0.00$*}&{$0.83$}&{$0.00$*} \\ 
\hspace{2mm}adj-dup &    &{$-0.10$}&{$0.00$*}&{$0.75$}&{$0.00$*} \\ 
\midrule
log weak-only ex. & $1.5$&{$-0.06$}&{$0.04$*}&{$-0.05$}&{$0.11$} \\ 
\hspace{2mm}pfx-dup &    &{$0.62$}&{$0.00$*}&{$0.19$}&{$0.00$*} \\ 
\hspace{2mm}first-last &    &{$0.58$}&{$0.00$*}&{$0.16$}&{$0.00$*} \\ 
\hspace{2mm}adj-dup &    &{$0.54$}&{$0.00$*}&{$-0.05$}&{$0.17$} \\ 
\midrule
log total ex. & $0.1$&{$-0.13$}&{$0.13$}&{$-0.09$}&{$0.26$} \\ 
I(label, weak) & $0.3$&{$-0.61$}&{$0.00$*}&{$-0.25$}&{$0.00$*} \\ 
$P(0|\neg weak)$ & $0.2$&{$-0.28$}&{$0.06$}&{$0.64$}&{$0.00$*} \\ 
$P(1|weak)$ & $0.2$&{$0.21$}&{$0.16$}&{$-0.61$}&{$0.00$*} \\ 
$P(0)$ & $0.1$&{$-0.44$}&{$0.01$*}&{$0.81$}&{$0.00$*} \\ 
\bottomrule
\end{tabular}
\caption{Regression results with terms for the interaction between feature type and number of examples added.}
\label{tab:mains-interactions}
\end{table*}
\begin{table*}[ht!]
\centering
\small
\setlength{\tabcolsep}{.5em}
\begin{tabular}{lcrlrlrlrl}
\toprule
&  & \multicolumn{2}{c}{\textit{weak-only} acc.} & \multicolumn{2}{c}{\textit{strong-only} acc.} & \multicolumn{2}{c}{\textit{both} acc.} & \multicolumn{2}{c}{\textit{neither} acc.} \\
&  & \multicolumn{2}{c}{$R^2=0.74$}&\multicolumn{2}{c}{$R^2=0.70$} & \multicolumn{2}{c}{$R^2=0.39$}&\multicolumn{2}{c}{$R^2=0.40$} \\
\cmidrule(lr){3-4} \cmidrule(lr){5-6}  \cmidrule(lr){7-8}  \cmidrule(lr){9-10}
& $\sigma$ & \multicolumn{1}{c}{$\beta$} & \multicolumn{1}{c}{$p$} & \multicolumn{1}{c}{$\beta$} & \multicolumn{1}{c}{$p$} 
& \multicolumn{1}{c}{$\beta$} & \multicolumn{1}{c}{$p$} & \multicolumn{1}{c}{$\beta$} & \multicolumn{1}{c}{$p$} \\
%\cmidrule(lr){2-2} \cmidrule(lr){3-4} \cmidrule(lr){5-6}
contains-1 &    &{$0.89$}&{$0.00$*}&{$0.90$}&{$0.00$*}&{$0.43$}&{$0.00$*}&{$0.49$}&{$0.00$*} \\ 
prefix-dupl &    &{$-0.07$}&{$0.03$*}&{$-0.01$}&{$0.71$}&{$0.17$}&{$0.00$*}&{$-0.14$}&{$0.00$*} \\ 
first-last &    &{$-0.26$}&{$0.00$*}&{$-0.43$}&{$0.00$*}&{$0.10$}&{$0.02$*}&{$0.03$}&{$0.52$} \\ 
adjacent-dupl &    &{$-0.58$}&{$0.00$*}&{$-0.48$}&{$0.00$*}&{$-0.73$}&{$0.00$*}&{$-0.39$}&{$0.00$*} \\ 
\midrule
log strong-only ex. & $1.5$&{$-0.14$}&{$0.00$*}&{$0.46$}&{$0.00$*}&{$0.28$}&{$0.00$*}&{$-0.72$}&{$0.00$*} \\ 
\midrule
log weak-only ex. & $1.5$&{$0.37$}&{$0.00$*}&{$0.03$}&{$0.25$}&{$-0.44$}&{$0.00$*}&{$-0.11$}&{$0.00$*} \\ 
\midrule
log total ex. & $0.1$&{$-0.15$}&{$0.16$}&{$-0.08$}&{$0.48$}&{$-0.32$}&{$0.05$}&{$-0.77$}&{$0.00$*} \\ 
I(label, weak) & $0.3$&{$-0.61$}&{$0.00$*}&{$-0.25$}&{$0.00$*}&{$-0.22$}&{$0.04$*}&{$-0.83$}&{$0.00$*} \\ 
$P(0|\neg weak)$ & $0.2$&{$-0.27$}&{$0.13$}&{$0.63$}&{$0.00$*}&{$0.22$}&{$0.42$}&{$0.79$}&{$0.00$*} \\ 
$P(1|weak)$ & $0.2$&{$0.22$}&{$0.22$}&{$-0.63$}&{$0.00$*}&{$0.29$}&{$0.28$}&{$0.22$}&{$0.40$} \\ 
$P(0)$ & $0.1$&{$-0.45$}&{$0.03$*}&{$0.82$}&{$0.00$*}&{$0.13$}&{$0.70$}&{$0.14$}&{$0.65$} \\ 
\bottomrule
\end{tabular}
\caption{Regression results for predicting all accuracy metrics. Fourth section are variables meant to control for confounding effects of adding counterexamples. $I()$ is mutual information between the presence of the weak feature and label according to training data; $P()$ are label skews according to training.}
\label{tab:mains-all-metrics}
\end{table*}

\end{document}